\documentclass{article}
\usepackage{spconf,amsmath,graphicx}
\usepackage{booktabs}
\usepackage{multirow}
\usepackage{amssymb}
\usepackage{pifont}
\usepackage{hyperref}

\usepackage{enumitem}
\setlist{nosep, leftmargin=14pt}

\usepackage{mwe} 

\title{TM-UNet: Token-Memory Enhanced Sequential Modeling for Efficient Medical Image Segmentation}
\name{Yaxuan Jiao$^{1,*}$, Qing Xu$^{2,3,5*}$, Yuxiang Luo$^{4*}$, Xiangjian He$^{5}$, Zhen Chen$^{6\dag}$, Wenting Duan$^{2}$\thanks{$^*$Equal contribution. $^\dag$Corresponding author.}}
\address{$^1$Dalian University of Technology, $^2$University of Lincoln, $^3$University of Nottingham, \\ $^4$Waseda University, $^5$University of Nottingham Ningbo China, $^6$Yale University}

\begin{document}
%
\maketitle
\begin{abstract}
Medical image segmentation is essential for clinical diagnosis and treatment planning. Although transformer-based methods have achieved remarkable results, their high computational cost hinders clinical deployment. To address this issue, we propose TM-UNet, a novel lightweight framework that integrates token sequence modeling with an efficient memory mechanism for efficient medical segmentation. Specifically, we introduce a multi-scale token-memory (MSTM) block that transforms 2D spatial features into token sequences through strategic spatial scanning, leveraging matrix memory cells to selectively retain and propagate discriminative contextual information across tokens. This novel token-memory mechanism acts as a dynamic knowledge store that captures long-range dependencies with linear complexity, enabling efficient global reasoning without redundant computation. Our MSTM block further incorporates exponential gating to identify token effectiveness and multi-scale contextual extraction via parallel pooling operations, enabling hierarchical representation learning without computational overhead. Extensive experiments demonstrate that TM-UNet outperforms state-of-the-art methods across diverse medical segmentation tasks with substantially reduced computation cost. The code is available at \url{https://github.com/xq141839/TM-UNet}.
\end{abstract}


\section{Introduction}
\label{sec:intro}

Medical image segmentation is fundamental to modern healthcare applications, enabling precise identification of anatomical structures and pathological regions for diagnostic assistance, surgical planning, and treatment evaluation \cite{ma2024segment}. Despite substantial advances, achieving robust and accurate segmentation remains challenging due to imaging noise, boundary ambiguity, and inter-patient anatomical variability across modalities.

The introduction of U-Net \cite{ronneberger2015u} marked a breakthrough in biomedical image segmentation with its symmetric encoder-decoder architecture and skip connections that effectively preserve multi-scale anatomical features \cite{chen2024transunet, chen2025sam}. Nevertheless, convolutional neural networks (CNNs) remain limited by their inherently local receptive fields, which restrict global context modeling and hinder precise delineation of complex lesions \cite{chen2024tinyu, chen2022aau}. To overcome this limitation, Vision Transformer (ViT) has emerged as a powerful alternative capable of capturing long-range spatial dependencies through self-attention mechanisms \cite{duan2025visionrwkv}. Their token-based representation allows global reasoning across entire medical images, substantially improving contextual understanding and segmentation performance.

\begin{figure}[!t]
  \centering
  \includegraphics[width=1\linewidth]{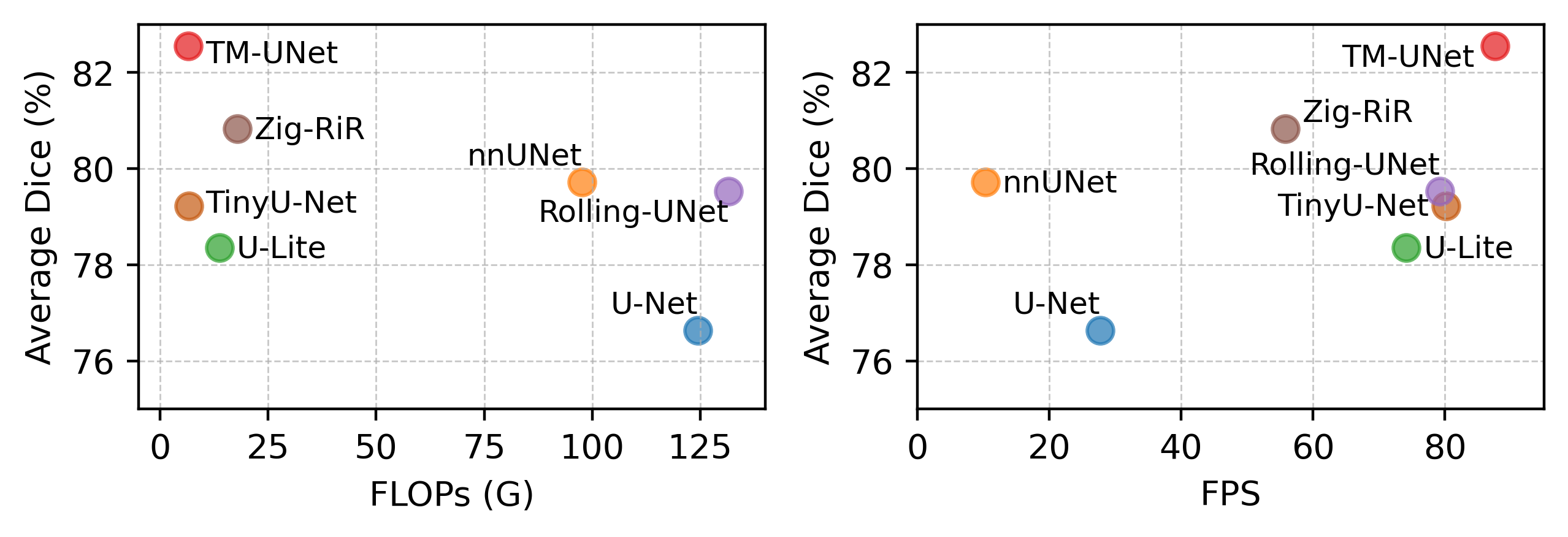}
  \caption{Performance and computation efficiency comparison of state-of-the-art medical image segmentation methods. Results demonstrate the superior performance of TM-UNet with lower computation complexity and faster inference speed.}
  \label{fig:intro}
\end{figure}

However, despite the progress made by existing medical ViT studies \cite{chen2024transunet, chen2025sam}, their practical deployment faces significant computational challenges that limit real-world clinical applications. The self-attention mechanism in ViT architectures introduces quadratic computation complexity with respect to the number of tokens, resulting in high memory usage and latency that are incompatible with real-time or resource-limited healthcare environments \cite{xu2024lb}. Moreover, many image tokens, often corresponding to non-pathological or background regions, contribute little discriminative information, leading to redundant computations and inefficiency. These challenges motivate the need for a computationally efficient mechanism that selectively focuses on informative tokens while maintaining the capacity to model long-range spatial dependencies.

\begin{figure*}[!t]
	\begin{center}
		\includegraphics[width=0.95\linewidth]{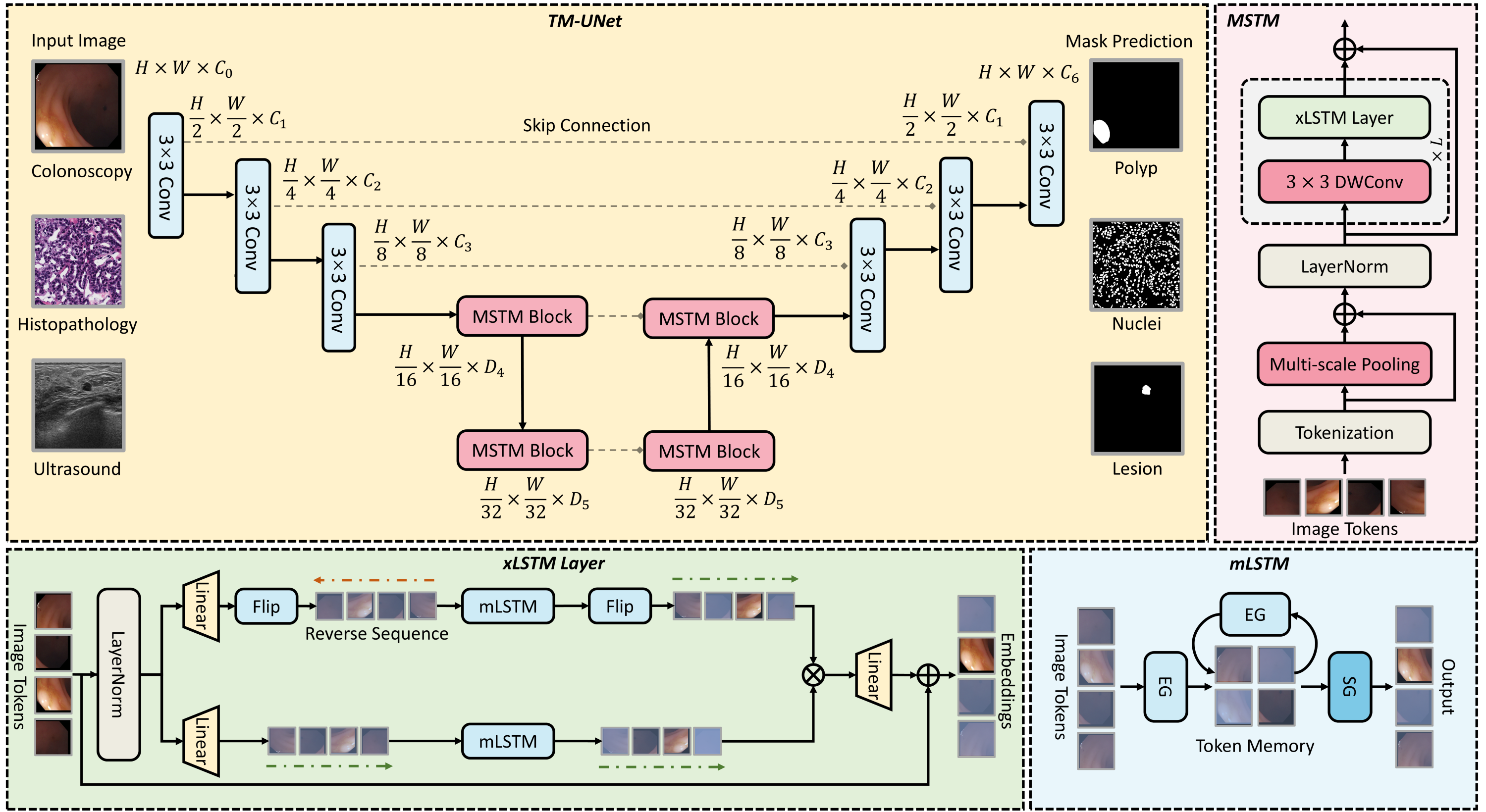}
	\end{center}
	\caption{The overview of our TM-UNet framework for efficient medical image segmentation. For ease of understanding, we elaborate on the
    case of TM-UNet in polyp segmentation. Our TM-UNet fully exploits token sequence modeling to achieve efficient long-range dependency modeling with linear computation complexity.}
	\label{fig:method}
\end{figure*}

To address this research gap, we introduce the token-memory concept, a compact and dynamic representation that enables tokens to access and update shared contextual information stored in memory cells. Unlike transformer architectures that repeatedly recompute pairwise relationships, token-memory accumulates and propagates salient contextual cues across tokens over time, reducing redundancy and preserving global coherence. Building on this concept, we propose TM-UNet, a novel framework that integrates token sequence modeling with a multi-scale token-memory (MSTM) block that transforms 2D spatial features into token sequences via strategic spatial scanning, leveraging matrix memory cells with covariance updates to capture long-range dependencies at linear computational complexity. An exponential gating mechanism further emphasizes discriminative tokens while filtering redundant information, and parallel pooling operations enable multi-scale contextual extraction without additional computational overhead. Extensive experiments across diverse segmentation tasks demonstrate that TM-UNet achieves state-of-the-art performance while substantially improving computational efficiency.

\begin{table*}[!t]
    \centering
    \small
    \setlength\tabcolsep{4.9pt}
    \caption{Comparison with state-of-the-arts on medical image segmentation.}
    {\scalebox{0.80}{
    \begin{tabular}{l|cc|cccc|cccc|cccc|cccc}
    \hline
    \multirow{2}{*}{Methods} & \multirow{2}{*}{FLOPs$\downarrow$} & \multirow{2}{*}{FPS$\uparrow$} & \multicolumn{4}{c|}{CVCCDB} & \multicolumn{4}{c|}{MoNuSeg} & \multicolumn{4}{c|}{ColonDB} & \multicolumn{4}{c}{UDIAT}\\
    \cline{4-19}
    & & & Dice$\uparrow$ & IoU$\uparrow$ & F1$\uparrow$ & HD$\downarrow$ & Dice$\uparrow$ & IoU$\uparrow$ & F1$\uparrow$ & HD$\downarrow$ & Dice$\uparrow$ & IoU$\uparrow$ & F1$\uparrow$ & HD$\downarrow$ & Dice$\uparrow$ & IoU$\uparrow$ & F1$\uparrow$ & HD$\downarrow$\\
    \hline
    U-Net \cite{ronneberger2015u} & 124.48G & 27.72 & 80.73 & 73.88 & 83.65 & 79.06 & 77.57 & 63.78 & 78.95 & 38.33 & 76.47 & 68.45 & 77.82 & 99.76 & 71.74 & 62.36 & 74.23 & 51.17\\
    nnUnet \cite{isensee2021nnu} & 97.62G & 10.35 & 84.72 & 77.45 & 86.93 & 58.73 & 81.97 & 69.49 & 82.13 & 33.95 & 77.97 & 70.86 & 79.54 & 87.42 & 74.21 & 63.74 & 76.89 & 48.35\\
    U-Lite \cite{dinh20231m} & 13.68G & 74.14 & 83.48 & 76.12 & 85.74 & 92.64 & 78.73 & 65.09 & 79.25 & 41.59 & 76.96 & 69.23 & 78.15 & 96.83 & 74.24 & 62.95 & 75.36 & 52.78\\ 
    TinyU-Net \cite{chen2024tinyu} & \underline{6.63G} & \underline{80.11} & 85.19 & 77.68 & 87.12 & \underline{48.22} & 79.72 & 66.84 & 80.73 & 34.99  & 76.87 & 69.65 & 78.44 & 91.25 & 75.12 & 64.08 & 77.85 & 46.92\\
    Rolling-Unet \cite{liu2024rolling} & 131.48G & 79.22 & 84.93 & 77.31 & 86.45 & 86.54 & 81.63 & 69.27 & 81.95 & 33.02 & 77.68 & 70.14 & 78.96 & 88.76 & 73.86 & 63.21 & 75.94 & 49.63\\
    Zig-RiR \cite{chen2025zig} & 17.89G & 55.81 & \underline{85.44} & \underline{78.01} & \underline{87.57} & 49.57 & \underline{83.71} & \underline{71.86} & \underline{83.42} & \underline{30.45} & \underline{78.06} & \underline{71.38} & \underline{79.87} & \underline{85.94} & \underline{76.12} & \underline{64.97} & \underline{78.53} & \underline{45.28}\\ 
    \hline
    TM-UNet & \textbf{6.55G} & \textbf{87.55} & \textbf{86.57} & \textbf{79.89} & \textbf{88.83} & \textbf{35.35} & \textbf{85.63} & \textbf{74.10} & \textbf{85.95} & \textbf{25.66} & \textbf{80.62} & \textbf{74.43} & \textbf{82.13} & \textbf{72.70} & \textbf{77.40} & \textbf{66.82} & \textbf{80.40} & \textbf{39.74}\\ 
    \hline
    \end{tabular}}}
    \label{tab_main_models}
\end{table*}

\section{Methodology}
\label{sec:format}

As illustrated in Fig. \ref{fig:method}, we present the TM-UNet framework for efficient medical image segmentation. To accomplish this, we design the multi-scale token-memory (MSTM) block to capture long-range dependencies at linear complexity. The proposed MSTM block employs Extended Long Short-Term Memory (xLSTM) \cite{alkin2025visionlstm} to identify token effectiveness with sequential modeling. This integrated design enables our TM-UNet framework to achieve superior segmentation performance with outstanding computation efficiency.

\subsection{Preliminaries}
The classical LSTM \cite{hochreiter1997long} was introduced to overcome the vanishing gradient problem of recurrent neural networks by controlling information flow through gating mechanisms:
\begin{equation}
c_t = f_t \odot c_{t-1} + i_t \odot z_t, \quad h_t = o_t \odot \psi(c_t),
\end{equation}
where $c_t$ is the internal cell state that accumulates long-term context from preceding steps. The forget, input, and output gates ($f_t$, $i_t$, and $o_t$) determine how much information from the previous memory $c_{t-1}$ and the current candidate $z_t$ is retained. The recent xLSTM \cite{beck2024xlstm} enhances this formulation through exponential gating, enabling more flexible and expressive information propagation. Its exponential input gate is defined as:
\begin{equation}
i_t = \exp(\tilde{i}_t), \quad \tilde{i}_t = \mathbf{w}_i^{\top}\mathbf{x}_t + r_i h_{t-1} + b_i,
\end{equation}
where $\mathbf{w}_i$, $r_i$ and $b_i$ are learnable parameters. This offers unbounded dynamic range for information acceptance, while the output gate maintains sigmoid activation $o_t = \sigma(\tilde{o}_t)$ for stability. The forget gate utilizes either sigmoid or exponential activation:
\begin{equation}
f_t = \sigma(\tilde{f}_t) \text{ or } f_t = \exp(\tilde{f}_t).
\end{equation}
This hybrid gating strategy enables the effective processing of sequential medical image features.

\begin{figure}[!t]
  \centering
  \includegraphics[width=0.95\linewidth]{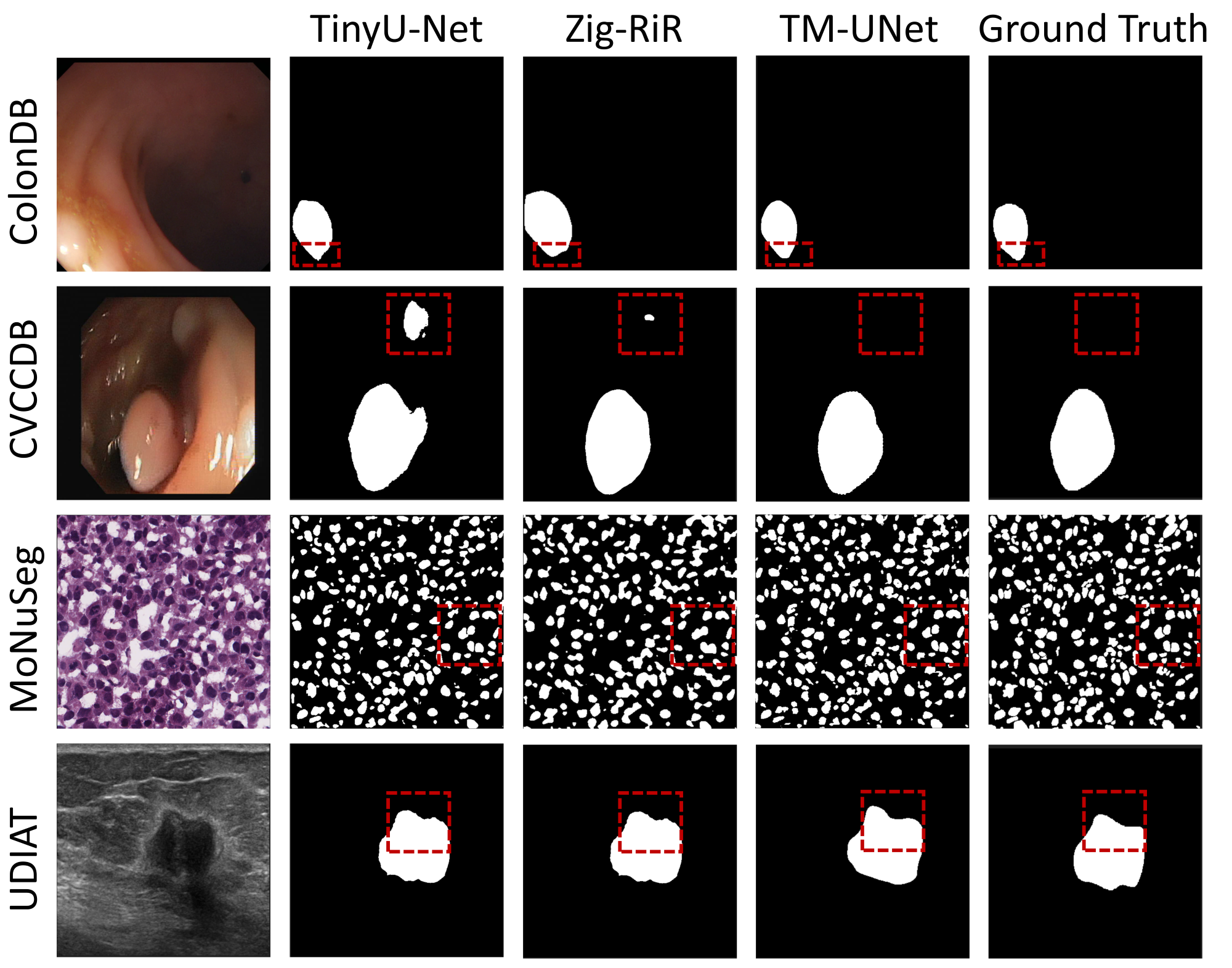}
  \caption{Qualitative comparison on four medical datasets.}
  \label{fig:intro}
\end{figure}

\subsection{Extended Long Short-Term Memory Layer}

To convert the formulation introduced in Section 2.1 into a practical network component for capturing long-range spatial dependencies, the bidirectional matrix LSTM (mLSTM) is employed to process through alternating scanning directions to ensure comprehensive spatial modeling. Given an input feature map $x$, the layer processes features through dual pathways with linear projections and flip operations. The bidirectional processing is formulated as:
\begin{equation}
x' = \text{Linear}(\text{LayerNorm}(x)),
\end{equation}
\begin{equation}
x'' = \text{Flip}(\text{mLSTM}(\text{Flip}(\text{Linear}(x)))),
\end{equation}
\begin{equation}
x \gets {\rm Linear}(x'' \odot \text{mLSTM}(x')) + x,
\end{equation}
where the flip operations reverse sequence order to enable bidirectional spatial modeling, and the final output combines both directional processing results through residual connections. The mLSTM computation provides flexible processing modes with different computational complexities. For sequence length $T$ and hidden dimension $d$, we employ the chunkwise mode with complexity $O(TSd + \frac{T}{S}d^2)$, where $S$ represents chunk size, enabling linear scaling suitable for practical medical image processing.

\subsection{Multi-Scale Token-Memory Block}

Multi-scale feature representation is critical for medical image segmentation, as anatomical structures and lesion regions vary widely in size and spatial distribution. Building on the xLSTM layer foundation, the proposed MSTM block implements the core token-memory principle, using compact memory cells to accumulate and reuse contextual information across spatial tokens. Specifically, a multi-scale pooling component, $f_{\rm pool}(\cdot)$, first captures differential contextual cues at multiple receptive-field scales without extra computation. The resulting representations are then processed through $L$ xLSTM layers, where each layer updates shared matrix-based memory cells that store salient token interactions, enabling adaptive long-range dependency modeling with linear complexity. Each xLSTM layer is followed by a $3 \times 3$ depthwise convolution to refine local spatial structure. The sequential pathway is formulated as:

\begin{equation}
h = \text{xLSTM}(\text{DWConv}(f_{\rm pool}(x)))),
\end{equation}
where $\text{DWConv}$ includes batch normalization and ReLU activation. This alternating structure balances computational efficiency and representation capability. The final embedding is computed in a residual form to stabilize training and preserve fine-grained detail: 

\begin{equation}
h \gets \text{LayerNorm}(f_{\text{pool}}(x)) + x+h.
\end{equation}
By combining multi-scale contextual awareness with sequential memory modeling, the MSTM block enables hierarchical understanding of medical images while maintaining the lightweight efficiency required for clinical deployment.

\subsection{The Optimization of TM-UNet Framework}

The complete TM-UNet integrates the proposed MSTM block within a U-Net framework optimized for efficient medical image segmentation. It adopts a hybrid encoder-decoder architecture that strategically places MSTM modules at deeper stages to capture long-range dependencies while maintaining computational efficiency. The encoder processes input images of size $H \times W \times C_0$ through five progressive downsampling stages. The first three stages employ conventional convolutional blocks with $3 \times 3$ convolutions, batch normalization, and ReLU activation to extract local anatomical features efficiently, producing feature maps at resolutions $\{C_1,C_2,C_3,C_4,C_5\}$. The final two stages incorporate MSTM modules to enhance global contextual reasoning.

The decoder mirrors the encoder with five upsampling stages. Each stage begins with bilinear upsampling followed by convolutional refinement, and integrates skip connections from corresponding encoder layers via channel-wise concatenation to preserve fine-grained spatial details. The two deepest decoder stages also include MSTM modules, ensuring consistent multi-scale context modeling and architectural symmetry. The network is optimized using a joint segmentation loss (cross-entropy and Dice losses) that balances pixel-wise accuracy and region-level consistency:
\begin{equation}
\mathcal{L}_{\rm seg}=\lambda_{1}\mathcal{L}_{\rm CE}+\lambda_{2}\mathcal{L}_{\rm Dice},
\end{equation}
where $\lambda_{1}$ and $\lambda_{2}$ are weighting factors. Through this hybrid design, TM-UNet combines the spatial inductive bias of convolutional networks with the global dependency modeling of MSTM modules, achieving accurate and efficient medical image segmentation with linear computational complexity.

\begin{table}[!t]
    \centering
    \small
    \setlength\tabcolsep{3.5pt}
    \caption{Ablation study of our TM-UNet framework.}
    {\scalebox{0.85}{
    \begin{tabular}{ccc|cc|cccc}
    \hline
    xLSTM & Pooling & DWConv & FLOPs$\downarrow$ & FPS$\uparrow$ & Dice$\uparrow$ & IoU$\uparrow$ & F1$\uparrow$ & HD$\downarrow$ \\
    \hline
     &  &  & \textbf{5.82G} & 89.15 & 82.47 & 70.31 & 81.68 & 29.87 \\
    \checkmark &  &  & 6.14G & \textbf{90.23} & 83.98 & 72.21 & 83.78 & 26.87 \\
    \checkmark & \checkmark &  & 6.14G & 88.92 & 84.82 & 73.45 & 84.36 & 26.33 \\
    \checkmark & \checkmark & \checkmark & 6.55G & 87.55 & \textbf{85.63} & \textbf{74.10} & \textbf{84.95} & \textbf{25.66} \\
    \hline
    \end{tabular}}}
    \label{tab_ablation_study}
\end{table}

\begin{table}[!t]
    \centering
    \small
    \setlength\tabcolsep{3.5pt}
    \caption{Analysis of TM-UNet with different channel sizes.}
    {\scalebox{0.88}{
    \begin{tabular}{ccccc|cc|cccc}
    \hline
    $C_1$ & $C_2$ & $C_3$ & $C_4$ & $C_5$ & FLOPs$\downarrow$ & FPS$\uparrow$ & Dice$\uparrow$ & IoU$\uparrow$ & F1$\uparrow$ & HD$\downarrow$ \\
    \hline
    8 & 16 & 32 & 64 & 128 & \textbf{0.49G} & \textbf{104.23} & 82.95 & 70.89 & 83.24 & 29.62 \\
    16 & 32 & 64 & 128 & 256 & 1.74G & 93.81 & 84.12 & 72.45 & 84.44 & 27.10 \\
    32 & 64 & 128 & 256 & 512 & 6.55G & 87.55 & \textbf{85.63} & \textbf{74.10} & \textbf{85.95} & \textbf{25.66} \\
    \hline
    \end{tabular}}}
    \label{tab_ulstm_variants}
\end{table}

\section{Experiments}

\subsection{Datasets and Implementation Details}

To validate the effectiveness of the proposed TM-UNet, we conduct comprehensive evaluations on four medical image datasets: CVCCDB \cite{bernal2015wm}, MoNuSeg \cite{kumar2017dataset}, ColonDB \cite{tajbakhsh2015automated}, and UDIAT \cite{yap2017automated}. We split all datasets into the training, validation, and test sets as 7:1:2, and all images are resized to $512 \times 512$ during the training and testing stage. We perform all experiments on a single NVIDIA A100 GPU using PyTorch. For fair comparisons, all segmentation methods are implemented with the same training settings and configurations. The loss coefficients $\lambda_1$ and $\lambda_2$ are set to 2 and 1, respectively. We apply the optimizer using Adam with an initial learning rate of $1\times10^{-4}$ and use the exponential decay strategy to adjust the learning rate with a factor of 0.98. The batch size and the training epoch are set to 16 and 200. 

\subsection{Comparison with State-of-the-art Methods}

To comprehensively evaluate the performance of TM-UNet, we conduct extensive comparisons with state-of-the-art methods, including classical U-Net architectures \cite{ronneberger2015u, isensee2021nnu}, lightweight models \cite{dinh20231m, chen2024tinyu}, and recent advanced approaches \cite{liu2024rolling, chen2025zig}. As shown in Table \ref{tab_main_models}, TM-UNet consistently outperforms the other methods across all four datasets while maintaining superior computational efficiency. Specifically, TM-UNet achieves 86.57\% Dice on CVCCDB (1.13\% improvement over high-performing Zig-RiR \cite{chen2025zig}), 85.63\% Dice on MoNuSeg (1.92\% improvement), 80.62\% Dice on ColonDB (2.56\% improvement), and 77.40\% Dice on UDIAT with the lowest HD of 39.74. Moreover, TM-UNet outperforms with only 6.55G FLOPs (1.30\% reduction compared to TinyU-Net \cite{chen2024tinyu}) and the highest FPS of 87.55, representing a 9.3\% improvement over the most efficient baseline. These results confirm that TM-UNet effectively addresses the long-standing trade-off between segmentation accuracy and computational cost, achieving SOTA performance with markedly improved efficiency suitable for practical clinical deployment.

\subsection{Ablation Study}

To investigate the effectiveness of MSTM components, we conduct comprehensive ablation studies in Table \ref{tab_ablation_study}. Starting from the baseline with a pure encoder-decoder architecture, we progressively integrate xLSTM, multi-scale pooling, and depth-wise convolution. The xLSTM improves Dice from 82.47\% to 83.98\%, validating its long-range dependency capture. Multi-scale pooling further enhances performance to 84.82\%, demonstrating effective contextual aggregation across scales. The complete TM-UNet achieves 85.63\% Dice and 74.10\% IoU, representing 3.16\% and 3.79\% improvements over baseline with marginal FLOPs increase. Table \ref{tab_ulstm_variants} analyzes channel configurations: the lightweight variant \{8, 16, 32, 64, 128\} achieves 82.95\% Dice with 0.49G FLOPs at 104.23 FPS for resource-constrained scenarios, while our standard configuration \{32, 64, 128, 256, 512\} attains optimal balance with 85.63\% Dice at 87.55 FPS. These results validate the contribution of xLSTM, pooling, and DWConv in MSTM for the superior performance of TM-UNet.

\section{Conclusion}

In this work, we present TM-UNet, an efficient medical image segmentation framework that balances accuracy and computational cost. By integrating U-Net with xLSTM through the proposed token-memory mechanism in the MSTM block, TM-UNet captures long-range dependencies with linear complexity. Extensive experiments demonstrate that TM-UNet outperforms state-of-the-art methods while achieving superior computation efficiency.

\bibliographystyle{IEEEbib}
\bibliography{refs}

\end{document}